\documentclass{article}

\usepackage[nonatbib, final]{tackling_climate_workshop_style}

\newif\ifarxiv\arxivtrue  
\ifarxiv
  \makeatletter
  \renewcommand{\@noticestring}{%
    40th Conference on Neural Information Processing Systems (NeurIPS 2026), Sydney, Australia.
    Workshop: Tackling Climate Change with Machine Learning.%
  }
  \makeatother
\fi

\usepackage[utf8]{inputenc} 
\usepackage[T1]{fontenc}    
\usepackage{hyperref}       
\usepackage{url}            
\usepackage{booktabs}       
\usepackage{multirow}       
\usepackage{longtable}      
\usepackage{float}          
\usepackage{amsmath}        
\usepackage{amsfonts}       
\usepackage{nicefrac}       
\usepackage{microtype}      
\usepackage{graphicx}       
\usepackage[style=numeric, sorting=none, backend=biber, natbib=true, maxcitenames=2]{biblatex} 
    \title{\textit{From Weather to Enhanced Weathering:} Weather Foundation Models for Carbon Dioxide Removal at Operational Scale}

\author{%
Sasankh Munukutla\thanks{Corresponding author.} \\
  Terradot Soil, Inc. \\
  San Francisco, CA, USA \\
  \texttt{sasankh@terradot.earth} \\ 
  \And
  Felipe Oviedo \\
  Microsoft AI for Good \\
  Redmond, WA, USA
   \And
  Danrong Zhang \\
  Microsoft Research \\
  Redmond, WA, USA
\And
  Kenji Takeda \\
  Microsoft Research \\
  Cambridge, UK
\And
  Juan M. Lavista Ferres \\
  Microsoft AI for Good \\
  Redmond, WA, USA
}

\begin{document}

\maketitle

\begin{abstract}
  Enhanced Rock Weathering (ERW) is a carbon removal solution with the potential to remove gigatons of CO$_2$ per year, but quantifying that removal today requires dense soil sampling and laboratory analysis. We evaluate five weather Foundation Models (FMs) as frozen encoders feeding lightweight decoders, under a leave-one-field-out protocol on 5,192 chronosequences from 264 fields across Brazil and the US. To our knowledge, this is the first holistic evaluation of weather FMs for ERW and the largest evaluation on operational ERW data. We find that across weather FMs, representations transfer to ERW: our best model, TerraNova-ERW, built on Aurora 1.5, positively orders measurements within fields it has never seen. Zero-shot prediction of a new field's mean removal is not yet feasible, and we attribute this to data scale: the same architecture excels on $\sim$$10^5$ natural-weathering sequences, anchoring the scale we expect ERW data must reach. Few-shot prediction, blending the model with a few samples per field, is supported today: TerraNova-ERW recovers over 60\% of the highest-removal fields with two samples per field, roughly halves the samples needed to estimate a field's mean removal, and can cut sampling on our densest fields by 28--98\%.
  \end{abstract}

\section{Introduction}

To avoid the most adverse effects of climate change, nearly every mitigation pathway requires Carbon Dioxide Removal (CDR) to balance hard-to-abate residual emissions, with $\sim$10 gigatons of CDR required annually by 2050 \cite{ipcc2022}. 

Enhanced Rock Weathering (ERW) is a promising CDR solution, which accelerates Earth's natural weathering process that draws down carbon over geological timescales. By applying finely crushed silicate rocks, such as basalt, to agricultural soils, ERW accelerates this process to occur in years rather than millennia. Scaled globally, ERW has the potential to remove gigatons of CO$_2$ per year while providing agronomic co-benefits such as improved crop yields and soil health \cite{beerling2020potential}.  

The key obstacle to scaling up ERW today is the Measurement, Reporting, and Verification (MRV) required to quantify the CDR. Quantifying removal currently requires dense physical ground-truth samples, including soil samples, along with rigorous laboratory analysis at every site. Process-based models such as Reactive Transport Models (RTMs) have been used to model ERW, but they require explicitly defining and approximating all relevant processes, such as mineral kinetics, and can be challenging to parametrize and computationally expensive to scale. Meanwhile, deep learning is increasingly applied across climate and Earth system problems \cite{nearing2024flood}, and weather Foundation Models (FMs) pretrained on decades of reanalysis data offer rich representations of the atmospheric drivers, such as precipitation, temperature, and soil moisture, that govern weathering rates. 

Previous work by \textcite{munukutla}, to our knowledge, the first application of deep learning to ERW, trained lightweight decoders on frozen representations from Microsoft's Aurora \cite{bodnar2024aurora} to show that atmospheric FM representations transfer to rock weathering. We build upon and advance this work by evaluating five weather FM backbones on our own operational ERW data spanning Brazil and the US, to our knowledge, the first holistic evaluation of multiple weather FMs for ERW and the largest evaluation on operational ERW data. We then show how our best model, TerraNova-ERW, can be used operationally to reduce the sampling MRV requires.

\section{Methods}

\subsection{Task and Dataset} We formulate ERW modeling as a sequence regression task using atmospheric FM representations: given monthly historic atmospheric representations and site-specific features  (static soil, dynamic site, amendment, spatial, temporal; see Appendix~\ref{app:features} for the full list of features) at $t_i$, predict at $t_{i+1}$ the CDR rate in t\,CO$_2$e/ha/yr.

Similar to \textcite{munukutla}, we assemble a dataset of natural weathering data using the same sources and process. We augment this with ERW data from our own deployments in the US and Brazil. The data is assembled as chronosequences, such that a given training example is a pair of measurements from two time points. In total, we have $192{,}511$ sequences of natural weathering data with global coverage and $5{,}192$ sequences of our
own operational ERW data spanning 264 agricultural fields. The two datasets play distinct roles: all main ERW results (Tables~\ref{tab:lofo} and~\ref{tab:harmonized}) are on the ERW
sequences. The natural-weathering data is used only in two separate
experiments: a joint-training experiment (Appendix \ref{app:full-results}) and a natural-only scale study (Appendix Table~\ref{tab:natural-weathering}). 

\subsection{Model Architecture}

The general architecture of TerraNova-ERW is summarized in Appendix~\ref{app:arch}; key elements follow.

\paragraph{Weather FM Backbones:} We consider five leading weather FMs to generate latents: Aurora 1.5 \cite{aurora1.5}, Pangu \cite{bi2023pangu}, SFNO \cite{SFNO}, FCN3 \cite{FCN3}, and GraphCast \cite{GraphCast}. For each backbone, we precompute global monthly latent representations by running the pretrained, frozen model on primarily ERA5 \cite{hersbach2020era5}, extracting the activations from the final encoder layer. At each site, we pool the latent field at three fixed spatial scales (local, regional, global) and concatenate them into a single $d_m$-dimensional embedding per month, obtained for every month spanning $t_i$ to $t_{i+1}$ of a chronosequence; this keeps training lightweight. Full details on the pooling scales and per-backbone extraction are in Appendix~\ref{app:backbone}. 

\paragraph{Temporal Aggregation:}The interval spanning a chronosequence can vary, and the $T$ monthly embeddings spanning $t_i$ to $t_{i+1}$ must be reduced to a fixed-size representation. We aggregate by trapezoidal integration of the latent trajectory, selected over supervised and self-supervised LTAE \cite{ltae} alternatives (Appendix~\ref{app:temporal}).

\paragraph{Lightweight Decoders:}
We build upon \textcite{lehmann2025finetuning} and \textcite{munukutla} to train lightweight decoders that fuse the aggregated weather representation with site features to predict the target. Considering the backbone latents are high-dimensional while labeled ERW data is scarce, decoder capacity is an important choice: we sweep the fusion mechanism and the per-backbone latent-projection width, favoring shallow decoders with wide hidden layers (``light wide''); the backbone latent is separately down-projected before fusion. Throughout, \emph{TerraNova-ERW} denotes the Aurora~1.5 backbone paired with a light-wide decoder and a $d{=}64$ latent projection (``64d''), with two refinements to its inputs and training: the three location inputs enter without per-fold standardization, and early stopping monitors validation RMSE rather than MAE. This is the configuration selected in Section~\ref{sec:results}. 

\subsection{Training and Evaluation}\label{sec:train-eval}

\paragraph{Leave-One-Field-Out (LOFO):} We train and evaluate all models by rotating the held-out field. Fields sharing a 2-decimal location key are merged into \emph{components} (264 fields $\to$ 101), since splitting fields that are spatially close would leak. Each fold trains with field-disjoint ERW validation, and predicts the held-out component (full LOFO protocol details in Appendix~\ref{app:lofo-details}). We favor this evaluation approach over a traditional held-out validation split given the dataset size.

\paragraph{Point vs.\ Site protocol:} \emph{Point Level} scores every held-out measurement, testing whether the model captures within-field structure: the capability \emph{model-and-measure (few-shot)}  MRV would use, where model predictions can help complement physical samples. \emph{Site Level} compares the predicted field mean to the true field mean, testing whether the model can place an unseen field's overall removal without any on-field measurement: the capability \emph{fully model-based (zero-shot)} MRV would require.

\paragraph{Metrics and Uncertainty:} We report MAE, RMSE, and Spearman rank correlation throughout; the row-shuffled ERA5 experiment serves as the pooled no-information reference. Uncertainty ($\pm$) is a bootstrap over held-out \emph{fields}: resampling the scored components (64 for treated-row metrics) with replacement 1000 times. At the point level, we also report ``\% best'', the share of the closable error a model removes:
\[
  \%\,\text{best} \;=\; 100 \times \frac{\text{constant}-\text{model}}{\text{constant}-\text{best}}
\]
where the constant is each fold's treated training mean and the best is each field's own measured median (MAE) or mean (RMSE), the best any single value per field can achieve. It is undefined at site level, where the best case is the exact field mean.

\paragraph{Experimental Configurations:} All experiments use trapezoidal aggregation and vary a single design choice, so each contrast isolates one effect: one configuration per weather FM backbone (which backbone transfers best), raw ERA5 inputs in place of the latent (the contribution of the FM encoding), and row-shuffled ERA5 (isolating the value of weather information itself), plus natural-weathering co-training, input-group ablation, and tabular AutoML configurations detailed in Appendix~\ref{app:full-results}. 

\section{Results}\label{sec:results}

\begin{table}[H]
  \centering
  \small
  \setlength{\tabcolsep}{3.5pt}
  \caption{LOFO results on treated rows. Point level scores every held-out measurement; site level compares predicted vs.\ true field means. \textbf{Bold} marks the best value in each block; ``\% best'' is defined in Section~\ref{sec:train-eval} and reported at point level only. Per-country (US, Brazil) point and site breakdowns, backbone comparisons, and other configurations are in Appendix Table~\ref{tab:harmonized}.}
  \label{tab:lofo}
  \resizebox{\textwidth}{!}{%
  \begin{tabular}{llccccc}
    \toprule
    & Experimental Configuration & MAE $\downarrow$ & RMSE $\downarrow$ & Spearman $\uparrow$ & \% best MAE $\uparrow$ & \% best RMSE $\uparrow$ \\
    \midrule
    \multirow{3}{*}{\shortstack[l]{All ERW\\(point, $n{=}4244$)}}
      & Aurora 1.5 (64d) (TerraNova-ERW) & \textbf{3.06} $\pm$0.20 & \textbf{4.20} & $\mathbf{+0.29}$ & \textbf{41} & \textbf{54} \\
      & Raw ERA5 (83d)                   & 3.21 $\pm$0.24 & 4.45 & $+0.22$ & 19 & 18 \\
      & Shuffled ERA5 (noise)            & 3.32 $\pm$0.32 & 4.67 & $-0.01$ & 2 & $-14$ \\
    \midrule
    \multirow{3}{*}{\shortstack[l]{All ERW\\(site, $n{=}64$)}}
      & Aurora 1.5 (64d) (TerraNova-ERW) & \textbf{1.78} $\pm$0.18 & \textbf{2.57} & $\mathbf{+0.26}$ & --- & --- \\
      & Raw ERA5 (83d)                   & 1.91 $\pm$0.24 & 2.69 & $+0.08$ & --- & --- \\
      & Shuffled ERA5 (noise)            & 1.96 $\pm$0.41 & 2.74 & $-0.00$ & --- & --- \\
    \bottomrule
  \end{tabular}%
  }
\end{table}

\paragraph{FM representations transfer to CDR prediction across architectures; Aurora performs best.} We see that the weather FM representations transfer to CDR across all backbones, while shuffling the weather away destroys ranking, showing weather \emph{information} is necessary (all-ERW point MAE 3.06--3.26 and Spearman $+0.12$ to $+0.29$ across backbones and Aurora variants, vs.\ 3.32 and $-0.01$ shuffled; raw ERA5 3.21 and $+0.22$; Appendix Table~\ref{tab:harmonized}). The FM encoding matches raw ERA5 on error size at the base configuration, while TerraNova-ERW more than doubles the share of closable error removed (41\% of MAE and 54\% of RMSE vs.\ 19\% and 18\% for raw ERA5; Table~\ref{tab:lofo}), and is more robust.
Raw ERA5 ranking collapses on one of three seeds (per-seed Spearman $-0.02$ to $+0.22$), whereas the Aurora (base, 64d, 512d), SFNO, FCN3, and GraphCast encodings hold $+0.16$ to $+0.28$ on every seed; only Pangu and the natural-weathering co-training variant share ERA5's instability (Appendix Table~\ref{tab:perseed}). Aurora 1.5 with a light-wide decoder, a $d{=}64$ latent projection, unstandardized location inputs, and RMSE early stopping achieved the best value on every all-ERW metric (point MAE 3.06, RMSE 4.20, Spearman $+0.29$; site MAE 1.78, RMSE 2.57, Spearman $+0.26$) while being stable on every seed, so we adopt it as our best model, which we call TerraNova-ERW. It was selected among about a dozen decoder and training variants evaluated on the same 68 folds.

\paragraph{Model-and-Measure MRV is supported; fully model-based MRV is not yet.} At the point level on treated rows, weather FMs positively order held-out measurements on fields they have never seen (Appendix Table~\ref{tab:harmonized}) with TerraNova-ERW's pooled Spearman $+0.29$ (Brazil $+0.40$), which is the capability model-and-measure MRV builds on. At the site level the model also orders unseen fields (pooled Spearman $+0.26$; US $+0.17$, Brazil $+0.37$), but its error on an unseen field's mean is only marginally below the no-information reference: pooled site MAE is 1.78 $\pm$0.18 vs.\ 1.96 for Shuffled ERA5, and on the US subset 1.50 $\pm$0.16 vs.\ 1.57 (Appendix Table~\ref{tab:harmonized}). Single fields with few measurements are high-variance targets, so ranking is recoverable where absolute measurement is not. Fully model-based quantification of unseen fields is therefore not yet supportable.

\paragraph{Natural Weathering Scale:} Trained and evaluated exclusively on natural weathering chronosequences, the same TerraNova-ERW architecture reaches validation MAE 0.014 and RMSE 0.017, roughly 90\% lower error than the deployable constant predictor (Appendix Table~\ref{tab:natural-weathering}). Given $\sim$10$^5$ diverse sequences, this model family excels at natural weathering prediction. This anchors our expectations for the needed data scale for fully model-based MRV for ERW. From our own scale-up and the growth of the ERW industry, we estimate this scale will be reached in a few years. 

\section{Operational Deployment of TerraNova-ERW}

\begin{figure}[H]
  \centering
  \includegraphics[width=\linewidth]{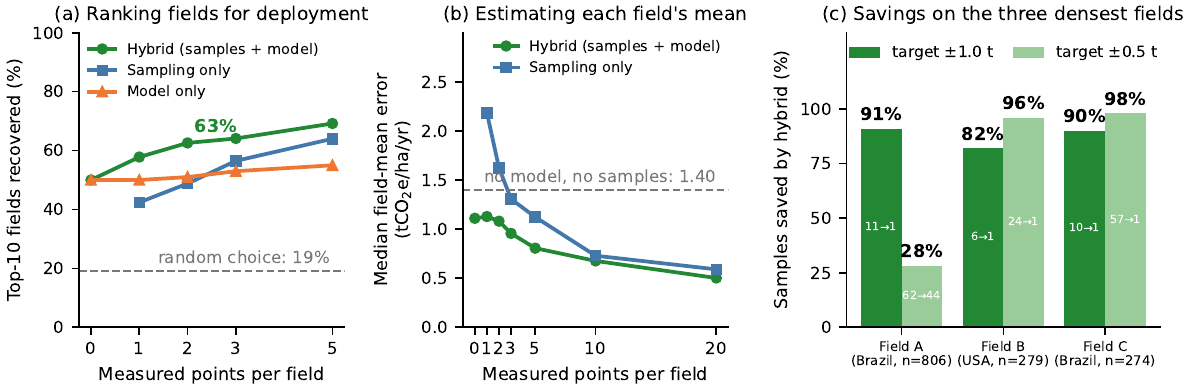}
  \caption{Sampling-budget simulation (Appendix~\ref{app:sampling}). (a) Share of the ten truly highest-removal fields recovered when ranking 52 fields under a budget of $k$ measured points per field. (b) Median error of the estimated field mean vs.\ budget. (c) Percent of samples the hybrid saves on the three densest fields at two precision targets.}
  \label{fig:sampling}
\end{figure}

\paragraph{Site Selection with Limited Features:} Site-specific features require physical samples taken at the field or, as a fallback, gridded products. We test whether a reduced-input variant of the model can operate before either is available, using only the weather FM latents, obtainable globally from a field's coordinates alone, together with the planned amendment specification. This Aurora $+$ Amendment configuration matches the full-input model at the same decoder settings on point-level error and orders unseen fields better (site Spearman $+0.32$ vs.\ $+0.13$ for Aurora 1.5 (base); Appendix~\ref{app:ablation}), and it remains competitive with TerraNova-ERW on site ranking ($+0.26$) without any site characterization. The advantage lies in ranking, not error: this supports prioritizing sites rather than quantifying them, and lets sampling and site visits be deferred.

\paragraph{Reducing Samples for MRV:} Since TerraNova-ERW supports model-and-measure MRV, we quantify what the model is worth in samples by simulating sparser campaigns on the LOFO out-of-fold predictions: each Monte-Carlo draw reveals $k$ random measured intervals per field, and a \emph{hybrid} estimator blends the revealed points with the model prediction at weights fitted out-of-sample (Protocol and results in Appendix~\ref{app:sampling}). Three results stand out (Figure~\ref{fig:sampling}). \emph{(1)} When ranking candidate fields, the hybrid recovers on average 63\% of the ten highest-removal fields with two samples per field, vs.\ 49\% for sampling alone at the same budget; the model alone, with no samples, already finds 50\%, as many as two samples per field (Figure~\ref{fig:sampling}a). \emph{(2)} At every budget below ${\sim}10$ points per field, the hybrid estimates field means with lower median error than sampling alone (1.13 vs.\ 2.18\,t\,CO$_2$e/ha/yr at one point), and the hybrid at three points matches what sampling alone needs five to ten points to reach (Figure~\ref{fig:sampling}b). \emph{(3)} On our three densest fields, which hold 32\% of all treated measurements, the hybrid saves 28--98\% of samples (Figure~\ref{fig:sampling}c): at the $\pm1.0$\,t target one sample replaces 6--11 samples on all three fields, and at $\pm0.5$\,t Fields~B and~C need a single sample (96\% and 98\% saved) because the model alone already places their means within 0.10 and 0.32\,t, while Field~A still needs 44 of 62 (28\% saved).

\section{Discussion}

\textbf{Pathway to climate impact:} MRV accounts for a third to over half of the cost of ERW \cite{mercer2024mrv}. Since sampling and laboratory analysis dominate MRV spend, cutting samples by 28--98\% attacks the largest term in a cost that can reach \$275/tCO$_2$ \cite{power2026erw}. TerraNova-ERW complements rather than replaces physical sampling, enabling sparser, targeted sampling designs and helping prioritize site selection. As latents can be precomputed globally and the decoder is lightweight, the marginal cost of modeling a new site is near zero, supporting cost-effective MRV as ERW scales to gigaton removal. 

\textbf{Limitations and future work:} Our deployment data covers only two countries and two feedstock types; temporal horizons are limited to monitoring to date; LOFO does not cover entirely new regions. Further validation across lithologies, climates, regions, and longer horizons along with training on more ERW data will enable models like TerraNova-ERW to be more robustly deployed. 

\newpage
\begin{ack} 
We gratefully acknowledge Katya Larina, Jackson Harris, Naomi Asimov Alley, Laisa de Assis Batista, Jono Finger, Jon Golla, Ella Holme, William Larsen, Jennifer Mills, Elaine Pegoraro, Silvia Palotti Polizel, Soroush Rezvan, and Adam Wolf for their invaluable contributions in assembling and curating the data.   We thank Prof.\ David Lobell and Prof.\ Scott Fendorf for their expert review and domain feedback, which significantly strengthened this work. We thank the anonymous reviewers for their constructive feedback. This work was supported by Terradot, which provided the operational ERW data. Finally, we thank the Microsoft AI for Good and Aurora teams for providing compute resources and for their generous support and responsiveness as we built on Aurora.
\end{ack}

\printbibliography

\appendix
\renewcommand{\thesubsection}{\Alph{section}\arabic{subsection}}
\section{Appendix}

\subsection{Architecture Diagram}\label{app:arch}

\begin{figure}[H]
  \centering
  \includegraphics[width=\linewidth]{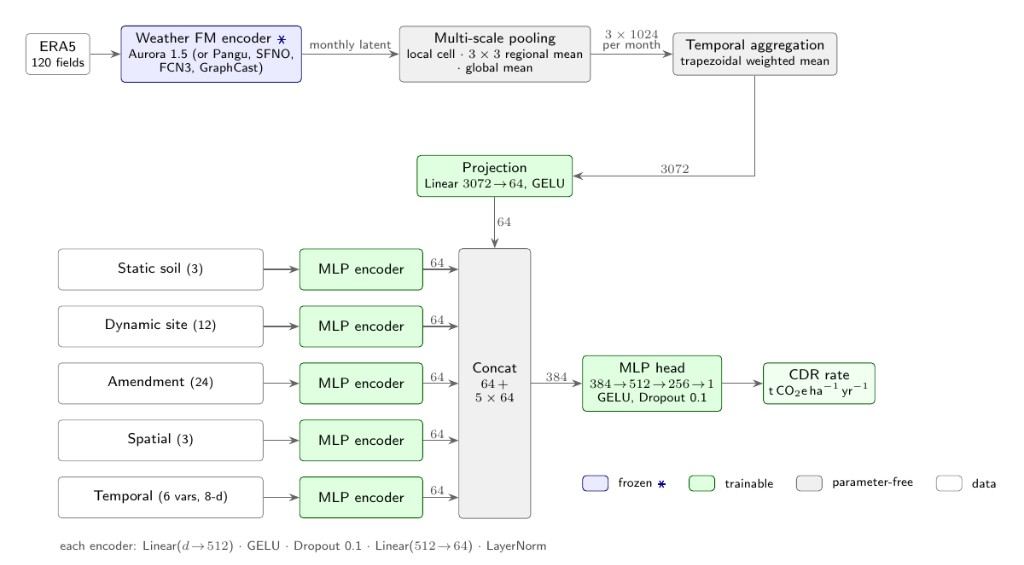}
  \caption{TerraNova-ERW architecture. A frozen weather foundation-model encoder (Aurora 1.5 by default; Pangu-Weather, SFNO, FCN3, and GraphCast are drop-in alternatives) maps ERA5 inputs to a global monthly latent, from which local (latent cell at the field location), regional ($3{\times}3$ mean), and global (map mean) features are pooled and aggregated over the observation window by a time-weighted trapezoidal mean. The aggregated 3072-d vector is projected to 64-d and concatenated with five 64-d MLP-encoded site-specific feature groups; a three-layer MLP head predicts the carbon dioxide removal (CDR) rate. Only the projection, site-specific feature encoders, and head are trained.}
  \label{fig:architecture}
\end{figure}

\subsection{Feature Variables}\label{app:features}

\begingroup
\setlength{\tabcolsep}{4pt}
\begin{longtable}{p{2cm}p{5.0cm}p{1.8cm}p{4.0cm}}
  \caption{Summary of input features and target variables used,
  organized by category. ERA5 variables are not consumed directly: they are fed
  to the frozen Aurora 1.5 encoder, and only its pooled
  backbone latent enters the predictor. The ERA5 block lists Aurora 1.5's
  complete input set: 18 surface fields plus a computed insolation channel, 36
  static fields, and 5 atmospheric variables on 13 pressure levels (50, 100,
  150, 200, 250, 300, 400, 500, 600, 700, 850, 925, 1000\,hPa), for 120 fields
  in total. Each field is supplied at two history timesteps (00:00 and 06:00 UTC
  on the first of the month). Parenthesized codes are the ERA5 / Aurora short
  names. If a given variable isn't available from the original dataset, it is obtained from an
  external source as shown below. Spatial and temporal variables are listed in
  physical units but are normalized before use (latitude by $90^\circ$,
  longitude to $[-1,1]$, area and interval by $\log(1+x)$, month and day of year
  by sine/cosine pairs).}
  \label{tab:feature-variables} \\
  \toprule
  \textbf{Category} & \textbf{Variable} & \textbf{Units} & \textbf{External Source} \\
  \midrule
  \endfirsthead

  \multicolumn{4}{c}{\textit{Table \ref{tab:feature-variables} continued from previous page}} \\
  \toprule
  \textbf{Category} & \textbf{Variable} & \textbf{Units} & \textbf{External Source} \\
  \midrule
  \endhead

  \midrule
  \multicolumn{4}{r}{\textit{Continued on next page}} \\
  \endfoot

  \bottomrule
  \endlastfoot

  ERA5 & \textit{Surface (18 fields\,+\,insolation):} & & \\
    & \quad 2m Temperature (2t) & K & ERA5 \citep{hersbach2020era5} \\
    & \quad 2m Dewpoint Temperature (2d) & K & ERA5 \\
    & \quad Skin Temperature (skt) & K & ERA5 \\
    & \quad Soil Temperature, level 1 (stl1) & K & ERA5 \\
    & \quad Volumetric Soil Water, level 1 (swvl1) & m$^3$/m$^3$ & ERA5 \\
    & \quad 10m U-wind (10u) & m/s & ERA5 \\
    & \quad 10m V-wind (10v) & m/s & ERA5 \\
    & \quad 100m U-wind (100u) & m/s & ERA5 \\
    & \quad 100m V-wind (100v) & m/s & ERA5 \\
    & \quad Mean Sea Level Pressure (msl) & Pa & ERA5 \\
    & \quad Surface Pressure (sp) & Pa & ERA5 \\
    & \quad Total Column Water Vapour (tcwv) & kg/m$^2$ & ERA5 \\
    & \quad Total Cloud Cover (tcc) & fraction & ERA5 \\
    & \quad Low Cloud Cover (lcc) & fraction & ERA5 \\
    & \quad Medium Cloud Cover (mcc) & fraction & ERA5 \\
    & \quad High Cloud Cover (hcc) & fraction & ERA5 \\
    & \quad Sea-Ice Cover (ci) & fraction & ERA5 \\
    & \quad Snow Depth (sd) & m w.e. & ERA5 \\
    & \quad Insolation & W/m$^2$ & Computed \cite{aurora1.5} \\
    & \textit{Static (36 fields):} & & \\
    & \quad Land-Sea Mask (lsm) & fraction & ERA5 \\
    & \quad Surface Geopotential (z) & m$^2$/s$^2$ & ERA5 \\
    & \quad Std.\ Dev.\ of Orography (sdor) & m & ERA5 \\
    & \quad Std.\ Dev.\ of Filtered Subgrid Orography (sdfor) & m & ERA5 \\
    & \quad Angle of Sub-gridscale Orography (anor) & rad & ERA5 \\
    & \quad Anisotropy of Sub-gridscale Orography (isor) & -- & ERA5 \\
    & \quad Slope of Sub-gridscale Orography (slor) & -- & ERA5 \\
    & \quad Lake Cover (cl) & fraction & ERA5 \\
    & \quad Lake Depth (dl) & m & ERA5 \\
    & \quad High Vegetation Cover (cvh) & fraction & ERA5 \\
    & \quad Low Vegetation Cover (cvl) & fraction & ERA5 \\
    & \quad Soil Type, one-hot: slt\_0--slt\_7 (8) & -- & ERA5 \\
    & \quad High Vegetation Type, one-hot: tvh\_\{0,3,4,5,6,18,19\} (7) & -- & ERA5 \\
    & \quad Low Vegetation Type, one-hot: tvl\_\{0,1,2,7,9,10,11,13,16,17\} (10) & -- & ERA5 \\
    & \textit{Atmospheric (5 variables $\times$ 13 levels):} & & \\
    & \quad Temperature (t) & K & ERA5 \\
    & \quad U-wind (u) & m/s & ERA5 \\
    & \quad V-wind (v) & m/s & ERA5 \\
    & \quad Specific Humidity (q) & kg/kg & ERA5 \\
    & \quad Geopotential (z) & m$^2$/s$^2$ & ERA5 \\
  \addlinespace
  \midrule
  Static Soil
    & Sand Content & \% & SoilGrids \citep{poggio2021soilgrids} \\
    & Silt Content & \% & SoilGrids \\
    & Clay Content & \% & SoilGrids \\
  \addlinespace
  \midrule
  Dynamic Site
    & Soil pH & pH & SoilGrids \\
    & Soil Cation Exchange Capacity & cmol/kg & SoilGrids \\
    & Soil Moisture & mm & GLDAS \citep{Rodell2004GLDAS} \\
    & Soil Runoff & mm/day & UNH/GRDC \citep{fekete2002runoff} \\
    & Evapotranspiration & mm/day & \citet{willmott2001et} \\
    & Soil Organic Carbon & g/kg & SoilGrids \\
    & Soil Erosion & kg/ha/day & GLOSEM \citep{borrelli2017glosem} \\
    & Soil Porosity & \% & GLDAS \\
    & Nitrification Rate & kg/ha/day & -- \\
    & Base Saturation & \% & HWSD \citep{FAO2023HWSDreport} \\
    & Net Primary Productivity & gC/m$^2$/day & MODIS \citep{zhao2005npp} \\
    & Precipitation & mm/year & ERA5 \\
  \addlinespace
  \midrule
  Amendment
    & \textit{Oxides:} & & \\
    & \quad CaO & \% & -- \\
    & \quad MgO & \% & -- \\
    & \quad Na$_2$O & \% & -- \\
    & \quad K$_2$O & \% & -- \\
    & \quad MnO & \% & -- \\
    & \quad SO$_3$ & \% & -- \\
    & \quad P$_2$O$_5$ & \% & -- \\
    & \textit{Minerals:} & & \\
    & \quad Quartz & \% & -- \\
    & \quad Calcite & \% & -- \\
    & \quad Actinolite & \% & -- \\
    & \quad Piemontite & \% & -- \\
    & \quad Chlorite & \% & -- \\
    & \quad Titanite & \% & -- \\
    & \quad Muscovite & \% & -- \\
    & \quad Albite & \% & -- \\
    & \quad Olivine & \% & -- \\
    & \quad Plagioclase & \% & -- \\
    & \quad Augite & \% & -- \\
    & \quad Orthopyroxene & \% & -- \\
    & \quad Magnetite & \% & -- \\
    & \quad Ilmenite & \% & -- \\
    & \quad Apatite & \% & -- \\
    & \textit{Physical:} & & \\
    & \quad Reactive Surface Area & m$^2$/g & Calculated if unavailable \\
    & \quad Application Rate & t/ha & Calculated if unavailable \\
  \addlinespace
  \midrule
  Spatial
    & Latitude & $^\circ$ & -- \\
    & Longitude & $^\circ$ & -- \\
    & Field Area & ha & -- \\
  \addlinespace
  \midrule
  Temporal
    & Observation Interval & days & -- \\
    & Month (sine/cosine) & -- & -- \\
    & Day of Year (sine/cosine) & -- & -- \\
    & Year & yr & -- \\
    & Amendment Age & days & -- \\
    & ERW Trial Indicator & binary & -- \\
  \addlinespace
  \midrule
  Target
    & CDR Rate & t\,CO$_2$e/ha/yr & ERW: measured$^*$; Natural Weathering: from source or calculated if unavailable \\
\end{longtable}
\endgroup

$^*$ The CDR rate labels for the operational ERW data are empirically derived and consider the near-field zone. Each georeferenced point is soil-sampled over a near-surface credit horizon at a pre-application baseline and at one or more later credit campaigns, and the samples undergo geochemical analysis. For each consecutive pair of samplings, removal is quantified by a solid-phase cation mass balance: an immobile trace element (e.g., Zr, Ti) constrains the rock mass present at the point, and the control-corrected loss of rock-derived cations is converted to CO$_2$ removal at 2 mol CO$_2$ per mol divalent cation, with corrections for background weathering, strong-acid (fertilizer-derived) weathering, and plant uptake. We use near-field CDR to denote the initial CO$_2$ uptake from cation release within the sampled soil horizon. Project emissions from quarrying, grinding, transport, and spreading are not deducted from the label: they do not vary at the sampling-point scale and are accounted for at issuance rather than in the spatial rate surface. Downstream loss and reversal factors are likewise not applied; unlike project emissions these do vary by catchment, so the labels are an upper bound on net durable removal. Labels are annualized by the interval length.

\begin{table}[H]
  \centering
  \small
  \setlength{\tabcolsep}{4pt}
  \caption{Frozen weather encoders implemented and evaluated, expressed relative
  to the Aurora 1.5 input set of Table \ref{tab:feature-variables}. Aurora 1.5 is
  the reference: Pangu-Weather, SFNO and FCN3 require strict subsets of its 120
  fields and are therefore served from the same cached ERA5 tensors by variable
  remap alone. GraphCast is the sole exception: it additionally requires vertical
  velocity ($w$) on all 13 pressure levels, which Aurora 1.5 does not use, so it
  is fetched separately from ARCO-ERA5 \citep{carver2023arco}. It also declares
  6-hourly total precipitation (tp06), but that field is target-only for the
  operational checkpoint and never a network input, so it is zero-filled.}
  \label{tab:encoder-inputs}
  \begin{tabular}{l p{4.3cm} p{6.0cm}}
    \toprule
    \textbf{Encoder} & \textbf{ERA5 input fields} & \textbf{Relation to Aurora 1.5 input set} \\
    \midrule
    Aurora 1.5
      & 18 surf.\ + insolation + 36 static + $5\times13$ atmos.\ (120)
      & Reference set \\
    \addlinespace
    Pangu-Weather
      & 4 surf.\ (2t, 10u, 10v, msl) + $5\times13$ atmos.\ (69)
      & Subset; drops all 36 static fields, insolation and 14 surface fields;
        levels reordered $1000\rightarrow50$ \\
    \addlinespace
    SFNO
      & 73 channels, $q$-based, 13 levels
      & Subset; no static fields or insolation \\
    \addlinespace
    FCN3
      & 72 channels, $q$-based, 13 levels
      & Subset; no static fields or insolation \\
    \addlinespace
    GraphCast
      & 4 surf.\ + 2 static (z, lsm) + $6\times13$ atmos.\ + tp06 (85);
        5 forcings computed internally
      & Not a subset: adds vertical velocity ($w$) on all 13 levels; tp06 is
        target-only and zero-filled \\
    \bottomrule
  \end{tabular}
\end{table}

\subsection{Backbone Latent Extraction Details}\label{app:backbone}

Each backbone is run pretrained and frozen on primarily ERA5, plus any additional data the given model requires (Table~\ref{tab:encoder-inputs}). At each site, we pool the extracted latent field at three fixed spatial scales and concatenate them: \emph{local}, the single latent cell containing the site (site-scale weather); \emph{regional}, the mean of the surrounding $3{\times}3$ window ($\sim$$\pm$112\,km; mesoscale context); and \emph{global}, the whole-map planetary scale; yielding a $3d_b = d_m$ dimensional embedding per month (e.g., $d_m{=}3072$ for Aurora 1.5's $d_b{=}1024$). Fixed scales are used rather than field-area-adaptive windows because fields are sub-cell at the latent grid resolution. Each backbone thus contributes only a $d_m$-dimensional per-month embedding (Aurora~1.5 = 3072, Pangu = 1152, GraphCast = 1536, SFNO = 1152, FCN3 = 168).

Table~\ref{tab:backbone-extraction} details, for each weather FM backbone, the layer from which latents are extracted, the native latent dimension $d_b$, and the resulting per-month embedding dimension $d_m$.

\begin{table}[H]
  \centering
  \small
  \caption{Latent extraction details for each weather FM backbone.}
  \label{tab:backbone-extraction}
  \begin{tabular}{lp{7.2cm}rr}
    \toprule
    Backbone & Layer extracted & $d_b$ & $d_m$ \\
    \midrule
    Aurora 1.5 & Output of \texttt{model.backbone} (3D Swin backbone, before the decoder); $d_b$ = \texttt{embed\_dim} $\times$ 2 = 1024 from the U-Net skip concat; 4 latent levels averaged & 1024 & 3072 \\
    \addlinespace
    Pangu-Weather & Features immediately before the \texttt{PatchRecovery} output layer; $d_b = 2 \times 192 = 384$ from the final upsample skip concat; 8 pseudo-levels (surface + 7 upper) averaged & 384 & 1152 \\
    \addlinespace
    SFNO & Input to \texttt{model.model.decoder}, i.e., output of the last SFNO block & 384 & 1152 \\
    \addlinespace
    FCN3 & Input to \texttt{model.model.decoder}, i.e., output of the last block before the decoder & 56 & 168 \\
    \addlinespace
    GraphCast & Grid-node features from the mesh2grid decoder's message passing, i.e., input to its final output MLP (\texttt{mesh2grid\_gnn.\_process} output, captured via haiku method interception); \texttt{latent\_size} = 512 & 512 & 1536 \\
    \bottomrule
  \end{tabular}
\end{table}

\subsection{Temporal Aggregation Methods}\label{app:temporal}

The $T$ monthly embeddings spanning $t_i$ to $t_{i+1}$, i.e.\ $\{z_1, \dots, z_T\}$, must be reduced to a fixed-size representation. We explore three methods:
\emph{(1) Trapezoidal Integration}, a fixed, parameter-free aggregation $\bar{z} = \frac{1}{T-1}\sum_{t=1}^{T-1} \frac{z_t + z_{t+1}}{2}$ approximating the time-integral of the latent trajectory, imposing the inductive bias that every month contributes;
\emph{(2) Supervised LTAE}, a Lightweight Temporal Attention Encoder \cite{ltae} trained end-to-end with the decoder to learn month-level importance weights (e.g., up-weighting wet seasons that dominate dissolution);
and \emph{(3) Self-supervised LTAE}, the same architecture trained with a masked-month reconstruction objective on unlabeled latent sequences, testing whether temporal structure can be learned without weathering-rate labels. 

Trapezoidal integration was selected for all final experiments after experimentation. 

\subsection{LOFO Protocol Details}\label{app:lofo-details}

Of the 101 location components, the 68 components with $\geq$10 measurements each receive a fold (5{,}062 of 5{,}192 rows). Each fold trains with field-disjoint ERW validation, early-stops on validation MAE (validation RMSE for TerraNova-ERW), ensembles seeds 42/43/44, and predicts the held-out component. Scalers are fit on training data only; TerraNova-ERW's three location inputs bypass the per-fold scaler. Four of the 68 components contain only control rows, so treated-row metrics cover 64 components (51 US, 13 Brazil). 

\subsection{Full Results}\label{app:full-results}

Each experimental configuration varies one design choice relative to the shared protocol, so each contrast isolates one effect: one configuration for each backbone (which weather FM backbone transfers best); joint training with natural weathering (the value of cross-regime pretraining); raw ERA5 inputs in place of the latent (the contribution of the FM encoding); row-shuffled ERA5 (real weather vectors, zero information, so the raw-vs-shuffled contrast isolates the value of weather information itself); input-group ablation, withholding the amendment and site-specific feature branches in turn (which inputs are actually required, and what is still recoverable when site characterization is unavailable; Appendix~\ref{app:ablation}); and AutoML (whether deep learning is needed). Folds, scalers, head, seeds, early stopping, and scoring are identical across configurations, except for TerraNova-ERW's two refinements (unstandardized location inputs and RMSE early stopping; Appendix~\ref{app:lofo-details}).

\begin{table}[H]
  \centering
  \scriptsize
  \setlength{\tabcolsep}{4pt}
  \renewcommand{\arraystretch}{0.92}
  \caption{Complete LOFO results on treated rows with a 1000-resample marginal bootstrap over held-out components on MAE. US pipeline: 55 components (51 with treated rows); Brazil pipeline: 13 components; site$=$field-mean prediction vs.\ field-mean truth. \textbf{Bold} marks the best value in each block. For Aurora 1.5, three variants are explored: the latents are projected to a narrower 64d and a wider 512d (note the latents are usually down-projected to 192d before fusing with site-specific features), and a variant with natural weathering co-training on 192,511 natural + $\sim$4,193 ERW rows per fold, evaluated on the identical ERW-only folds. The 64d row is TerraNova-ERW, which additionally uses unstandardized location inputs and RMSE early stopping (Appendix~\ref{app:lofo-details}). FLAML \cite{wang2021flaml}: AutoML run with all site-specific features. Search over GBM and tree-ensemble families; ExtraTrees was selected most often (32/68 folds), followed by LightGBM (21/68).
}
  \label{tab:harmonized}
  \begin{tabular}{llccc}
    \toprule
    & Model & MAE $\downarrow$ & RMSE $\downarrow$ & Spearman $\uparrow$ \\
    \midrule
    \multirow{11}{*}{\shortstack[l]{All ERW\\point\\$n{=}4244$}}
      & Aurora 1.5 (base)            & 3.21 $\pm$0.23 & 4.44 & $+0.24$ \\
      & Aurora 1.5 $+$ natural weathering co-training  & 3.26 $\pm$0.32 & 4.55 & $+0.12$ \\
      & Aurora 1.5 (64d) (TerraNova-ERW)                     & \textbf{3.06} $\pm$0.20 & \textbf{4.20} & $\mathbf{+0.29}$ \\
      & Aurora 1.5 (512d)                    & 3.22 $\pm$0.24 & 4.39 & $+0.25$ \\
      & Pangu-Weather (base)                    & 3.22 $\pm$0.25 & 4.49 & $+0.19$ \\
      & SFNO (base)                             & 3.23 $\pm$0.24 & 4.47 & $+0.19$ \\
      & FCN3 (base)                             & 3.19 $\pm$0.22 & 4.39 & $+0.21$ \\
      & GraphCast (base)                        & 3.25 $\pm$0.24 & 4.45 & $+0.21$ \\
      & Raw ERA5 (83d)                          & 3.21 $\pm$0.24 & 4.45 & $+0.22$ \\
      & Shuffled ERA5 (noise)                   & 3.32 $\pm$0.32 & 4.67 & $-0.01$ \\
      & FLAML (AutoML)                          & 3.25 $\pm$0.29 & 4.48 & $+0.22$ \\
    \midrule
    \multirow{11}{*}{\shortstack[l]{All ERW\\site\\$n{=}64$}}
      & Aurora 1.5 (base)                       & 2.04 $\pm$0.18 & 2.78 & $+0.13$ \\
      & Aurora 1.5 $+$ natural weathering co-training & 1.99 $\pm$0.37 & 2.71 & $+0.08$ \\
      & Aurora 1.5 (64d) (TerraNova-ERW)        & \textbf{1.78} $\pm$0.18 & \textbf{2.57} & $\mathbf{+0.26}$ \\
      & Aurora 1.5 (512d)                       & 1.99 $\pm$0.24 & 2.80 & $+0.15$ \\
      & Pangu-Weather (base)                    & 2.07 $\pm$0.21 & 2.81 & $+0.04$ \\
      & SFNO (base)                             & 2.03 $\pm$0.17 & 2.81 & $+0.04$ \\
      & FCN3 (base)                             & 2.04 $\pm$0.24 & 2.88 & $+0.01$ \\
      & GraphCast (base)                        & 1.95 $\pm$0.19 & 2.77 & $+0.08$ \\
      & Raw ERA5 (83d)                          & 1.91 $\pm$0.24 & 2.69 & $+0.08$ \\
      & Shuffled ERA5 (noise)                   & 1.96 $\pm$0.41 & 2.74 & $-0.00$ \\
      & FLAML (AutoML)                          & 2.02 $\pm$0.34 & 2.61 & $+0.07$ \\
    \midrule
    \multirow{11}{*}{\shortstack[l]{US\\point\\$n{=}2701$}}
      & Aurora 1.5 (base)                       & 2.67 $\pm$0.17 & 3.50 & $+0.16$ \\
      & Aurora 1.5 $+$ natural weathering co-training & \textbf{2.60} $\pm$0.18 & \textbf{3.43} & $\mathbf{+0.22}$ \\
      & Aurora 1.5 (64d) (TerraNova-ERW)        & 2.61 $\pm$0.18 & 3.46 & $+0.16$ \\
      & Aurora 1.5 (512d)                       & 2.68 $\pm$0.17 & 3.53 & $+0.15$ \\
      & Pangu-Weather (base)                    & 2.64 $\pm$0.20 & 3.53 & $+0.10$ \\
      & SFNO (base)                             & 2.69 $\pm$0.18 & 3.54 & $+0.10$ \\
      & FCN3 (base)                             & 2.66 $\pm$0.18 & 3.52 & $+0.07$ \\
      & GraphCast (base)                        & 2.70 $\pm$0.17 & 3.53 & $+0.10$ \\
      & Raw ERA5 (83d)                          & 2.65 $\pm$0.19 & 3.51 & $+0.12$ \\
      & Shuffled ERA5 (noise)                   & 2.63 $\pm$0.19 & 3.50 & $+0.11$ \\
      & FLAML (AutoML)                          & 2.73 $\pm$0.22 & 3.59 & $\mathbf{+0.22}$ \\
    \midrule
    \multirow{11}{*}{\shortstack[l]{US\\site\\$n{=}51$}}
      & Aurora 1.5 (base)                       & 1.69 $\pm$0.19 & 2.27 & $+0.10$ \\
      & Aurora 1.5 $+$ natural weathering co-training & 1.61 $\pm$0.19 & 2.15 & $+0.15$ \\
      & Aurora 1.5 (64d) (TerraNova-ERW)        & \textbf{1.50} $\pm$0.16 & \textbf{2.13} & $\mathbf{+0.17}$ \\
      & Aurora 1.5 (512d)                       & 1.68 $\pm$0.17 & 2.32 & $+0.14$ \\
      & Pangu-Weather (base)                    & 1.74 $\pm$0.18 & 2.36 & $-0.01$ \\
      & SFNO (base)                             & 1.72 $\pm$0.19 & 2.33 & $-0.03$ \\
      & FCN3 (base)                             & 1.68 $\pm$0.18 & 2.31 & $-0.10$ \\
      & GraphCast (base)                        & 1.64 $\pm$0.18 & 2.27 & $+0.05$ \\
      & Raw ERA5 (83d)                          & 1.60 $\pm$0.19 & 2.25 & $+0.08$ \\
      & Shuffled ERA5 (noise)                   & 1.57 $\pm$0.18 & 2.22 & $-0.01$ \\
      & FLAML (AutoML)                          & 1.86 $\pm$0.27 & 2.40 & $-0.04$ \\
    \midrule
    \multirow{11}{*}{\shortstack[l]{Brazil\\point\\$n{=}1543$}}
      & Aurora 1.5 (base)                       & 4.17 $\pm$0.38 & 5.71 & $+0.26$ \\
      & Aurora 1.5 $+$ natural weathering co-training & 4.41 $\pm$0.48 & 6.03 & $+0.03$ \\
      & Aurora 1.5 (64d) (TerraNova-ERW)        & \textbf{3.86} $\pm$0.39 & \textbf{5.25} & $\mathbf{+0.40}$ \\
      & Aurora 1.5 (512d)                       & 4.16 $\pm$0.36 & 5.60 & $+0.29$ \\
      & Pangu-Weather (base)                    & 4.25 $\pm$0.38 & 5.79 & $+0.12$ \\
      & SFNO (base)                             & 4.19 $\pm$0.34 & 5.73 & $+0.14$ \\
      & FCN3 (base)                             & 4.11 $\pm$0.40 & 5.60 & $+0.29$ \\
      & GraphCast (base)                        & 4.21 $\pm$0.36 & 5.71 & $+0.20$ \\
      & Raw ERA5 (83d)                          & 4.19 $\pm$0.37 & 5.73 & $+0.22$ \\
      & Shuffled ERA5 (noise)                   & 4.54 $\pm$0.42 & 6.21 & $-0.08$ \\
      & FLAML (AutoML)                          & 4.17 $\pm$0.43 & 5.71 & $+0.25$ \\
    \midrule
    \multirow{11}{*}{\shortstack[l]{Brazil\\site\\$n{=}13$}}
      & Aurora 1.5 (base)                       & 3.40 $\pm$0.55 & 4.21 & $+0.11$ \\
      & Aurora 1.5 $+$ natural weathering co-training & 3.48 $\pm$0.76 & 4.26 & $-0.19$ \\
      & Aurora 1.5 (64d) (TerraNova-ERW)        & 2.87 $\pm$0.54 & 3.83 & $\mathbf{+0.37}$ \\
      & Aurora 1.5 (512d)                       & 3.20 $\pm$0.75 & 4.17 & $+0.25$ \\
      & Pangu-Weather (base)                    & 3.36 $\pm$0.50 & 4.12 & $+0.09$ \\
      & SFNO (base)                             & 3.25 $\pm$0.55 & 4.20 & $+0.18$ \\
      & FCN3 (base)                             & 3.48 $\pm$0.81 & 4.45 & $+0.26$ \\
      & GraphCast (base)                        & 3.18 $\pm$0.61 & 4.18 & $+0.23$ \\
      & Raw ERA5 (83d)                          & 3.10 $\pm$0.55 & 3.96 & $+0.06$ \\
      & Shuffled ERA5 (noise)                   & 3.46 $\pm$0.83 & 4.21 & $-0.04$ \\
      & FLAML (AutoML)                          & \textbf{2.67} $\pm$0.64 & \textbf{3.31} & $+0.20$ \\
    \bottomrule
  \end{tabular}
\end{table}

Table~\ref{tab:perseed} breaks the
all-ERW treated point-level results out per seed. The FM encodings differ from raw ERA5 less in average performance than in reliability: raw ERA5 ranking collapses on one of
its three seeds (Spearman $-0.02$ on seed 43), and Pangu shows the same
failure mode on two seeds, while the Aurora (base, 64d, 512d), SFNO, FCN3, and GraphCast encodings stay
within a narrow band ($+0.16$ to $+0.28$) on all three. Shuffled ERA5 sits stably near
zero, as expected for the no-information reference. Since an operational MRV model is retrained as data accrues, a configuration whose ranking skill depends on the training seed is not deployable even if its ensemble looks competitive.

\begin{table}[H]
  \centering
  \small
  \setlength{\tabcolsep}{3pt}
  \caption{Per-seed pooled results, all-ERW treated rows, point level ($n{=}4244$ per
  seed; seeds 42/43/44). MAE in t\,CO$_2$e/ha/yr; $\rho$ is the Spearman rank
  correlation and sd($\rho$) is its standard deviation
  across the three seeds. \textbf{Bold} marks the best value in each column; \underline{underline} marks seed-level ranking collapses
  ($\rho \le 0.05$). FLAML is omitted.}
  \label{tab:perseed}
  \begin{tabular}{lcccccccc}
    \toprule
    & \multicolumn{2}{c}{seed 42} & \multicolumn{2}{c}{seed 43} & \multicolumn{2}{c}{seed 44} & \\
    \cmidrule(lr){2-3}\cmidrule(lr){4-5}\cmidrule(lr){6-7}
    Model & MAE & Spearman & MAE & Spearman & MAE & Spearman & sd($\rho$) \\
    \midrule
    Aurora 1.5 (base)                    & 3.20 & $+0.24$ & 3.20 & $+0.25$ & 3.30 & $+0.19$ & 0.03 \\
    Aurora 1.5 $+$ natural weathering co-training & 3.24 & $+0.17$ & 3.23 & $+0.15$ & 3.41 & \underline{$+0.05$} & 0.06 \\
    Aurora 1.5 (64d) (TerraNova-ERW)     & \textbf{3.09} & $\mathbf{+0.28}$ & 3.22 & $+0.22$ & \textbf{3.10} & $\mathbf{+0.28}$ & 0.03 \\
    Aurora 1.5 (512d)                    & 3.26 & $+0.23$ & 3.18 & $\mathbf{+0.26}$ & 3.27 & $+0.25$ & \textbf{0.01} \\
    Pangu-Weather (base)                 & 3.32 & \underline{$+0.03$} & 3.23 & $+0.22$ & 3.37 & \underline{$-0.04$} & 0.13 \\
    SFNO (base)                          & 3.27 & $+0.16$ & 3.27 & $+0.18$ & 3.21 & $+0.22$ & 0.03 \\
    FCN3 (base)                          & 3.23 & $+0.19$ & \textbf{3.16} & $+0.23$ & 3.20 & $+0.21$ & 0.02 \\
    GraphCast (base)                     & 3.24 & $+0.21$ & 3.32 & $+0.22$ & 3.29 & $+0.17$ & 0.03 \\
    Raw ERA5 (83d)                       & 3.20 & $+0.21$ & 3.32 & \underline{$-0.02$} & 3.22 & $+0.22$ & 0.14 \\
    Shuffled ERA5 (noise)                & 3.34 & \underline{$-0.03$} & 3.33 & \underline{$+0.00$} & 3.35 & \underline{$-0.00$} & 0.02 \\
    \bottomrule
  \end{tabular}
\end{table}

\begin{table}[H]
  \centering
  \small
  \setlength{\tabcolsep}{6pt}
  \caption{Natural-weathering validation performance of the same TerraNova-ERW architecture (64d decoder head on the frozen Aurora 1.5 encoder), trained and evaluated exclusively on natural weathering chronosequences: 173,070 training sequences and a location-disjoint validation set of 19,441 sequences, with no ERW data seen during training or validation. Values are the mean over five seeds; MAE and RMSE are stable across seeds ($\pm 0.001$). The target (CDR, t\,CO$_2$e/ha/yr) spans  0.185 to 0.686 over the training sequences and 0.211 to 0.681 over validation, with validation mean 0.254 and standard deviation $\sigma=0.081$. The deployable constant predictor, the training-set mean of 0.375, gives the reference row; against it the model scores $\sim$90\% lower MAE and RMSE. This run uses a similar configuration as the other experiments (standardized location inputs, MAE early stopping).}
  \label{tab:natural-weathering}
  \begin{tabular}{lccc}
    \toprule
    \textbf{Model} & \textbf{MAE} & \textbf{RMSE} & \textbf{Spearman $\rho$} \\
    \midrule
    Aurora 1.5 (64d) (TerraNova-ERW)
                   & \textbf{0.014} & \textbf{0.017} & 0.741 \\
    \textit{Constant predictor (training mean, 0.375)}
                   & 0.142 & 0.145 & --- \\
    \bottomrule
  \end{tabular}
\end{table}

\subsection{Input Ablation: Site Selection with Limited Features}\label{app:ablation}

We rerun the identical 68-fold LOFO protocol over four input configurations, holding folds, decoder, early stopping, and scoring fixed. Evaluation covers operational US and Brazil ERW rows only (5{,}062 rows; 4{,}244 treated, 818 control); the configurations are Aurora latents alone (no amendment and site-specific features), Amendment only (24 variables), Aurora $+$ Amendment, and Aurora $+$ full Site Specific Features. All four configurations are the Aurora 1.5 (base) configuration, so the full-input corresponds to the Aurora 1.5 (base) row of Appendix Table~\ref{tab:harmonized}, not TerraNova-ERW; TerraNova-ERW is included in Table~\ref{tab:ablation} as a reference row for comparison.
\begin{table}[H]
  \centering
  \small
  \setlength{\tabcolsep}{4pt}
  \caption{Input ablation on treated rows, same 68 LOFO folds as Table~\ref{tab:lofo}. MAE and RMSE in t\,CO$_2$e/ha/yr. Site metrics average predictions and targets within each held-out component before scoring.}
  \label{tab:ablation}
  \begin{tabular}{lcccccc}
    \toprule
    & \multicolumn{3}{c}{Point ($n{=}4244$)} & \multicolumn{3}{c}{Site ($n{=}64$)} \\
    \cmidrule(lr){2-4}\cmidrule(lr){5-7}
    Inputs & MAE $\downarrow$ & RMSE $\downarrow$ & Spearman $\uparrow$ & MAE $\downarrow$ & RMSE $\downarrow$ & Spearman $\uparrow$ \\
    \midrule
    Aurora 1.5 latent only          & 3.34 & 4.66 & $-0.12$ & \textbf{1.80} & 2.52 & $+0.08$ \\
    Amendment only & 3.26 & 4.62 & $+0.02$ & 2.01 & 2.77 & $+0.06$ \\
    Aurora $+$ Amendment        & 3.27 & 4.63 & $+0.11$ & \textbf{1.80} & \textbf{2.51} & $\mathbf{+0.32}$ \\
    Aurora $+$ full Site Specific Features & \textbf{3.21} & \textbf{4.44} & $\mathbf{+0.24}$ & 2.04 & 2.78 & $+0.13$ \\
    \addlinespace
    \emph{Ref: TerraNova-ERW} & 3.06 & 4.20 & $+0.29$ & 1.78 & 2.57 & $+0.26$ \\ 

  \bottomrule
  \end{tabular}
\end{table}

Aurora latents alone are not deployable: point Spearman is $-0.12$, so the model fails to order measurements within an unseen field, even though its site-level error is competitive. Adding the amendment variables is the single change that restores rank signal at both levels, lifting point Spearman to $+0.11$ and site Spearman to $+0.32$. Relative to the full stack, Aurora $+$ Amendment gives up little at the point level (MAE 3.27 vs.\ 3.21, Spearman $+0.11$ vs.\ $+0.24$) while ordering held-out fields far better (site Spearman $+0.32$ vs.\ $+0.13$, site MAE 1.80 vs.\ 2.04); a plausible reading is that soil and spatial covariates are informative about within-field variation but are partly absorbed as field identity, which does not transfer to an unseen component. One caveat: the site-level advantage is concentrated in ranking rather than error, with site MAE staying near 1.8--2.0\,t\,CO$_2$e/ha/yr across configurations, so these results support site prioritization rather than model-based quantification.

\subsection{Sampling-Budget Simulation}\label{app:sampling}

\paragraph{Protocol.} A \emph{field} is one location component and a \emph{point} is one measured CDR interval on that field. Each Monte-Carlo draw reveals $k$ randomly chosen points per field, forms an estimate of the field's mean, and scores it against the mean of the points that were \emph{not} revealed, so no estimate is graded on its own inputs. Model predictions are the LOFO out-of-fold predictions of TerraNova-ERW (Aurora~1.5, 64d latent projection) (68 folds, seeds 42--44 averaged); priors and variance components are likewise fitted leaving the scored field out. Estimators are: \emph{sampling only} ($\bar{y}_k$); \emph{sampling shrunk}, the $k$-point mean shrunk toward the leave-one-field-out prior with the error-optimal weight $w = \tau^2/(\tau^2 + \sigma^2/k)$ from the between- and within-field variance components, but with no model, isolating the value of shrinkage itself; \emph{model only} ($\hat{f}$ on the unrevealed points); \emph{hybrid (error weights)}, the same shrinkage applied to the model's residuals, i.e.\ the error-optimal blend of measurements and model; and \emph{hybrid (rank weights)}, $\alpha\, z(\bar{y}_k) + (1-\alpha)\, z(\hat{f})$ with one $\alpha$ per budget tuned for the ranking objective on the even-numbered Monte-Carlo draws, with every estimator scored on the odd-numbered draws only ($\alpha$ rises from 0.65 at one point to 0.85 at five). Ranking fixes the portfolio at the 52 fields with $\geq13$ treated points so every budget orders the same set; error tiers use every field with enough unrevealed points left to grade, so $n$ falls with $k$ (64 fields at $k{=}1$, 63 at $k{=}2$, 62 at $k{=}3$, 52 at $k{=}5$, 39 at $k{=}10$, 25 at $k{=}20$) and the fields surviving at high $k$ are systematically the denser ones.

Three caveats bound these numbers. First, the truth is itself an estimate (the mean of the unrevealed points), and within-field noise carries 64\% of total variance, which puts a floor under every error and caps attainable rank agreement. Second, rows at different budgets are not computed over identical field sets (see the \emph{fields} column of Table~\ref{tab:tier-budget}). Third, the residual ``poor'' tail is where measurements matter most: with no measured points the model alone leaves 13\% of fields in the ``poor'' tier (error ${>}4$\,t), and a single measured point per field brings that to 5\%, because even one point of local calibration removes a stable per-field offset in the model.

\begin{table}[H]
  \centering
  \small
  \caption{Ranking 52 fields for deployment potential under a measurement budget of $k$ points per field (mean over 100 held-out Monte-Carlo draws, $\pm$ across draws). ``Top-10 found'' is the share of the 10 genuinely highest-removal fields appearing in the estimator's top 10; picking at random gives 19\%. $\alpha$ is the tuned weight on the measured mean. The rank-weighted hybrid leads on Spearman and leads or ties on top-10 at every budget; the error-weighted hybrid is strongest on the top-quartile screen.}
  \label{tab:rank-budget}
  \begin{tabular}{lrcccc}
    \toprule
    Estimator & points & Spearman $\rho$ & top-10 found & top-quartile hit & $\alpha$ \\
    \midrule
    \multirow{4}{*}{Sampling only}
      & 1 & 0.443 $\pm$0.114 & 42\% & 49\% & --- \\
      & 2 & 0.529 $\pm$0.094 & 49\% & 57\% & --- \\
      & 3 & 0.599 $\pm$0.092 & 57\% & 63\% & --- \\
      & 5 & 0.661 $\pm$0.068 & 64\% & 69\% & --- \\
    \midrule
    \multirow{4}{*}{Sampling shrunk (no model)}
      & 1 & 0.402 $\pm$0.119 & 40\% & 46\% & --- \\
      & 2 & 0.504 $\pm$0.096 & 47\% & 55\% & --- \\
      & 3 & 0.580 $\pm$0.096 & 55\% & 62\% & --- \\
      & 5 & 0.649 $\pm$0.070 & 62\% & 69\% & --- \\
    \midrule
    \multirow{5}{*}{Model only}
      & 0 & 0.280            & 50\% & 54\% & --- \\
      & 1 & 0.286 $\pm$0.015 & 50\% & 53\% & --- \\
      & 2 & 0.292 $\pm$0.022 & 51\% & 56\% & --- \\
      & 3 & 0.289 $\pm$0.025 & 53\% & 54\% & --- \\
      & 5 & 0.299 $\pm$0.032 & 55\% & 54\% & --- \\
    \midrule
    \multirow{5}{*}{Hybrid (error weights)}
      & 0 & 0.245            & 40\% & 54\% & --- \\
      & 1 & 0.412 $\pm$0.074 & 56\% & 60\% & --- \\
      & 2 & 0.501 $\pm$0.073 & 62\% & 66\% & --- \\
      & 3 & 0.570 $\pm$0.065 & 64\% & 71\% & --- \\
      & 5 & 0.658 $\pm$0.059 & 69\% & 75\% & --- \\
    \midrule
    \multirow{5}{*}{Hybrid (rank weights)}
      & 0 & 0.280            & 50\% & 54\% & 0.00 \\
      & 1 & 0.476 $\pm$0.101 & 58\% & 60\% & 0.65 \\
      & 2 & 0.557 $\pm$0.086 & 63\% & 65\% & 0.70 \\
      & 3 & 0.617 $\pm$0.079 & 64\% & 69\% & 0.75 \\
      & 5 & 0.674 $\pm$0.064 & 69\% & 73\% & 0.85 \\
    \bottomrule
  \end{tabular}
\end{table}

\begin{table}[H]
  \centering
  \small
  \caption{Share of fields by accuracy of their estimated mean CDR under a budget of $k$ points per field: typical $|$error$|$ within 1, 1--2, 2--4, and beyond 4\,t\,CO$_2$e/ha/yr, against field means spanning $-5$ to $+10$. The no-model, no-point baseline is the $k{=}0$ row of ``sampling shrunk'', at 41\% great. The hybrid beats sampling alone at every budget through ten points and matches or beats shrinkage without a model through three points (shrinkage overtakes on the ``great'' tier at $k{=}5$, 69\% vs.\ 63\%, while the hybrid retains the lower median error through ten points). What measurements add to the hybrid is mainly the tail, with ``poor'' falling from 13\% at $k{=}0$ to 5\% with a single point.}
  \label{tab:tier-budget}
  \begin{tabular}{lrccccr}
    \toprule
    Estimator & points & great $\leq1$ & good 1--2 & fair 2--4 & poor $>4$ & fields \\
    \midrule
    \multirow{6}{*}{Sampling only}
      & 1  & 9\%  & 36\% & 41\% & 14\% & 64 \\
      & 2  & 16\% & 56\% & 24\% & 5\%  & 63 \\
      & 3  & 29\% & 52\% & 19\% & 0\%  & 62 \\
      & 5  & 42\% & 52\% & 6\%  & 0\%  & 52 \\
      & 10 & 72\% & 23\% & 5\%  & 0\%  & 39 \\
      & 20 & 96\% & 4\%  & 0\%  & 0\%  & 25 \\
    \midrule
    \multirow{7}{*}{Sampling shrunk (no model)}
      & 0  & 41\% & 22\% & 27\% & 11\% & 64 \\
      & 1  & 39\% & 31\% & 25\% & 5\%  & 64 \\
      & 2  & 43\% & 25\% & 27\% & 5\%  & 63 \\
      & 3  & 48\% & 32\% & 18\% & 2\%  & 62 \\
      & 5  & 69\% & 25\% & 6\%  & 0\%  & 52 \\
      & 10 & 77\% & 18\% & 5\%  & 0\%  & 39 \\
      & 20 & 92\% & 8\%  & 0\%  & 0\%  & 25 \\
    \midrule
    \multirow{7}{*}{Model only}
      & 0  & 45\% & 22\% & 20\% & 13\% & 64 \\
      & 1  & 44\% & 23\% & 20\% & 13\% & 64 \\
      & 2  & 46\% & 22\% & 19\% & 13\% & 63 \\
      & 3  & 47\% & 21\% & 21\% & 11\% & 62 \\
      & 5  & 48\% & 23\% & 21\% & 8\%  & 52 \\
      & 10 & 46\% & 31\% & 13\% & 10\% & 39 \\
      & 20 & 52\% & 36\% & 4\%  & 8\%  & 25 \\
    \midrule
    \multirow{7}{*}{Hybrid (error weights)}
      & 0  & 44\% & 25\% & 19\% & 13\% & 64 \\
      & 1  & 45\% & 30\% & 20\% & 5\%  & 64 \\
      & 2  & 48\% & 29\% & 19\% & 5\%  & 63 \\
      & 3  & 52\% & 27\% & 18\% & 3\%  & 62 \\
      & 5  & 63\% & 27\% & 8\%  & 2\%  & 52 \\
      & 10 & 77\% & 18\% & 5\%  & 0\%  & 39 \\
      & 20 & 92\% & 8\%  & 0\%  & 0\%  & 25 \\
    \bottomrule
  \end{tabular}
\end{table}

\begin{table}[H]
  \centering
  \small
  \caption{Samples required to estimate the field mean within a precision target on the three densest fields, which together hold 32\% of all treated points. Values are interpolated from per-field Monte-Carlo error curves (300 draws per budget): samples needed by sampling alone vs.\ the hybrid, and the resulting share of samples saved. On Fields~B and~C the $\pm0.5$\,t target is met with one sample because the model alone already places their means within 0.10 and 0.32\,t; this rests on two fields and should be read per field rather than as a typical saving. Field names are anonymized.}
  \label{tab:dense-savings}
  \setlength{\tabcolsep}{5pt}
  \begin{tabular}{llrcccc}
    \toprule
    & & & \multicolumn{2}{c}{Target $\pm1.0$\,t} & \multicolumn{2}{c}{Target $\pm0.5$\,t} \\
    \cmidrule(lr){4-5}\cmidrule(lr){6-7}
    Field & Country & points & sampling $\rightarrow$ hybrid & saved & sampling $\rightarrow$ hybrid & saved \\
    \midrule
    Field A & Brazil & 806 & 11 $\rightarrow$ 1 & 91\% & 62 $\rightarrow$ 44 & 28\% \\
    Field B & USA    & 279 & 6 $\rightarrow$ 1  & 82\% & 24 $\rightarrow$ 1 & 96\% \\
    Field C & Brazil & 274 & 10 $\rightarrow$ 1 & 90\% & 57 $\rightarrow$ 1 & 98\% \\
    \bottomrule
  \end{tabular}
\end{table}
 
\end{document}